\documentclass[sigconf,dvipsnames,prologue,table,xcdraw,]{acmart}
\AtBeginDocument{%
  \providecommand\BibTeX{{%
    \normalfont B\kern-0.5em{\scshape i\kern-0.25em b}\kern-0.8em\TeX}}}

\copyrightyear{2024}
\acmYear{2024}
\setcopyright{rightsretained}
\acmConference[SIGGRAPH Conference Papers '24]{Special Interest Group on Computer Graphics and Interactive Techniques Conference Conference Papers '24}{July 27-August 1, 2024}{Denver, CO, USA}
\acmBooktitle{Special Interest Group on Computer Graphics and Interactive Techniques Conference Conference Papers '24 (SIGGRAPH Conference Papers '24), July 27-August 1, 2024, Denver, CO, USA}
\acmDOI{10.1145/3641519.3657470}
\acmISBN{979-8-4007-0525-0/24/07}

\usepackage{cleveref}
\usepackage{graphicx}
\usepackage{lscape}
\usepackage{multirow}
\usepackage{pifont}
\usepackage{subcaption}
\usepackage{wrapfig}
\usepackage{xspace}

\newtoggle{arxiv}

\definecolor{darkred}{RGB}{204, 0, 0}
\definecolor{darkgreen}{RGB}{0, 160, 0}
\definecolor{darkorange}{rgb}{1.0, 0.55, 0.0}
\definecolor{darkorange}{rgb}{1.0, 0.55, 0.0}
\definecolor{darkgreen}{rgb}{0.0, 0.4, 0.0}

\creflabelformat{equation}{#2\textup{#1}#3}
\crefname{equation}{Eq.}{Eqs.}
\crefname{figure}{Fig.}{Figs.}
\crefname{section}{Sec.}{Secs.}
\crefname{table}{Tab.}{Tabs.}

\newcommand{\old}[1]{}

\newcommand{\greencheck}{\color{PineGreen}\ding{51}}
\newcommand{\crossred}{\color{Mahogany}\ding{55}}
\newcommand{\crossorange}{\color{CarnationPink}\ding{55}}

\newcommand{\setN}{\mathbb{N}}
\newcommand{\setR}{\mathbb{R}}

\newcommand{\OurMethod}{\textcolor{black}{Lite2Relight}~}

\newcommand{\gen}{\mathcal{G}_{sg}}
\newcommand{\decoder}{\mathcal{G}_{dec}}
\newcommand{\encoder}{\mathcal{E}}
\newcommand{\afa}{\emph{AFA}}

\newcommand{\latentNet}{\mathcal{R}}

\newcommand{\pose}{c} 
\newcommand{\inImg}{I} 
\newcommand{\latw}{{w}^+} 
\newcommand{\genFeat}{G^k}

\newcommand{\diffImg}{\Delta{\inImg}} 
\newcommand{\invImg}{I_{w^+}} 
\newcommand{\recFeat}{F} 

\newcommand{\imageolat}{\mathbf{O}} 
\newcommand{\natImg}{\mathbf{I}} 

\newcommand{\srcImg}{I_{s}} 
\newcommand{\srclatw}{\latw_{s}} 
\newcommand{\srcgenFeat}{\genFeat_{s}}
\newcommand{\srcrecFeat}{\recFeat_{s}} 

\newcommand{\emap}{E} 
\newcommand{\srcemap}{\emap_{s}} 
\newcommand{\tgtemap}{\emap_{t}} 
\newcommand{\tgtImg}{I_{t}} 
\newcommand{\tgtlatw}{\latw_{t}} 

\newcommand{\relitlatw}{\hat{w}^+_{t}} 

\newcommand{\deltalatw}{\Delta{w}} 
\newcommand{\relitgenFeat}{G^k_{r}}
\newcommand{\relitrecFeat}{\recFeat_{t}} 
\newcommand{\relitrecImg}{\hat{\inImg_{t}}} %

\newcommand{\loss}{\mathcal{L}}
\newcommand{\lossall}{\mathcal{L}_{\text{total}}}
\newcommand{\lossc}{\loss_\text{C}}
\newcommand{\lossl}{\loss_\text{lat}}
\newcommand{\lossp}{\loss_\text{LPIPS}}

\newcommand{\wtl}{\lambda_0}
\newcommand{\wtc}{\lambda_1}
\newcommand{\wtp}{\lambda_2}

\def\rk{{\textnormal{k}}}

\newcommand{\incid}{i} 
\newcommand{\factor}{E} 

\newcommand{\Norms}[1]{\left\| #1 \right\|_1} 
\newcommand{\Norm}[1]{\left\| #1 \right\|_2^2} 

\begin{document}
\title{Lite2Relight:  3D-aware Single Image Portrait Relighting }

\author{Pramod Rao}
\email{prao@mpi-inf.mpg.de}
\orcid{0009-0003-7236-169X}
\affiliation{%
  \institution{MPI for Informatics, SIC \& \\ VIA Research Center}
  \city{Saarbrücken}
  \country{Germany}
}

\author{Gereon Fox}
\email{gfox@mpi-inf.mpg.de}
\orcid{0009-0002-3471-7715}
\affiliation{%
  \institution{MPI for Informatics}
  \city{Saarbrücken}
  \country{Germany}
}

\author{Abhimitra Meka}
\email{abhim@google.com}
\orcid{0000-0001-7906-4004}
\affiliation{
  \institution{Google Inc.}
  \city{San Fransisco}
  \country{USA}
}

\author{ Mallikarjun B R}
\email{ mbr@mpi-inf.mpg.de}
\orcid{0009-0007-5906-8666}
\affiliation{%
  \institution{MPI for Informatics}
  \city{Saarbrücken}
  \country{Germany}
}
\author{Fangneng Zhan}
\email{fzhan@mpi-inf.mpg.de}
\orcid{0000-0003-1502-6847}
\affiliation{%
  \institution{MPI for Informatics}
  \city{Saarbrücken}
  \country{Germany}
}

\author{Tim Weyrich}
\email{tim.weyrich@fau.de}
\orcid{0000-0002-4322-8844}
\affiliation{%
  \institution{Friedrich-Alexander-Universität Erlangen-Nürnberg (FAU)}
  \city{Nürnberg}
  \country{Germany}
}

\author{Bernd Bickel}
\email{bernd.bickel@ist.ac.at}
\orcid{0000-0001-6511-9385}
\affiliation{%
  \institution{IST Austria}
  \city{Klosterneuburg}
  \country{Austria} \\
  \institution{ETH Zürich}
  \city{Zürich}
  \country{Switzerland}
}
\author{Hanspeter Pfister}
\email{pfister@g.harvard.edu}
\orcid{0000-0002-3620-2582}
\affiliation{%
  \institution{Harvard University}
  \city{Cambridge}
  \country{USA}
}

\author{Wojciech Matusik}
\email{wojciech@csail.mit.edu}
\orcid{0000-0003-0212-5643}
\affiliation{%
  \institution{MIT}
  \city{Cambridge}
  \country{USA}
}

\author{Mohamed Elgharib}
\email{elgharib@mpi-inf.mpg.de}
\orcid{0000-0001-8727-0895}
\affiliation{%
  \institution{MPI for Informatics}
  \city{Saarbrücken}
  \country{Germany}
}

\author{Christian Theobalt}
\email{theobalt@mpi-inf.mpg.de}
\orcid{0000-0001-6104-6625}
\affiliation{%
  \institution{MPI for Informatics, SIC \& \\ VIA Research Center}
  \city{Saarbrücken}
  \country{Germany}
}

\begin{abstract}
Achieving photorealistic 3D view synthesis and relighting of human portraits is pivotal for advancing AR/VR applications. 
Existing methodologies in portrait relighting demonstrate substantial limitations in terms of generalization and 3D consistency, coupled with inaccuracies in physically realistic lighting and identity preservation. 
Furthermore, personalization from a single view is difficult to achieve and often requires multiview images during the testing phase or involves slow optimization processes.
This paper introduces \OurMethod, a novel technique that can predict 3D consistent head poses of portraits while performing physically plausible light editing at interactive speed. 
Our method uniquely extends the generative capabilities and efficient volumetric representation of EG3D, leveraging a lightstage dataset to implicitly disentangle face reflectance and perform relighting under target HDRI environment maps. By utilizing a pre-trained geometry-aware encoder and a feature alignment module, we map input images into a relightable 3D space, enhancing them with a strong face geometry and reflectance prior.
%
Through extensive quantitative and qualitative evaluations, we show that our method outperforms the state-of-the-art methods in terms of efficacy, photorealism, and practical application. This includes producing 3D-consistent results of the full head, including hair, eyes, and expressions. \OurMethod paves the way for large-scale adoption of photorealistic portrait editing in various domains, offering a robust, interactive solution to a previously constrained problem. Project page: \textcolor{CarnationPink}{\url{https://vcai.mpi-inf.mpg.de/projects/Lite2Relight//}}.

\end{abstract}

\begin{teaserfigure}
  \includegraphics[width=\textwidth]{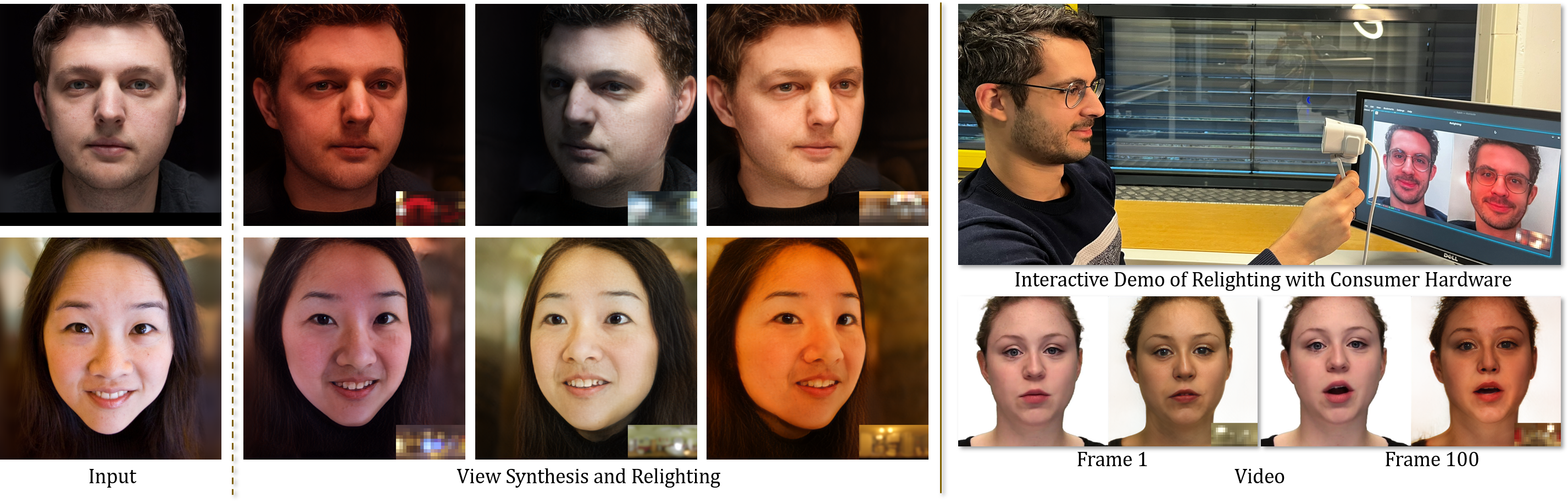}
  \caption{We present Lite2Relight, a method that can relight monocular portrait images given HDRI environment maps. Our method demonstrates strong generalization to in-the-wild images, maintains 3D consistent pose synthesis of the subjects and performs physically accurate relighting. Moreover, courtesy of our lightweight architecture, \OurMethod can relight subjects captured by a live webcam at interactive rates. Image credits to Flickr.}
  \Description{This is the teaser figure for the article. }
  \label{fig:teaser}
\end{teaserfigure}

\renewcommand\shortauthors{Rao et al.}

\ccsdesc[500]{Computing methodologies~Image representations}

\maketitle
\section{Introduction}
Photorealistic editing and compositing of human portrait images is a technical challenge underlying various graphics applications: computationally enhanced photography~\cite{sipr_sun}, content generation~\cite{Fried:2020:ES} and immersive telepresence~\cite{iphoneavatar} are some examples. Large-scale adoption of such applications is limited by the challenge of solving this interactively with minimal computing power using sparse data input, usually an image or video from a single camera. The underlying technical challenge consists of modeling the very diverse range of 3D geometry and reflectance of human heads from such sparse data and achieving a perceivable high degree of photorealism. 

An interactive photorealistic 3D portrait editing and relighting solution that can generalize to unseen subjects from a single in-the-wild input image does not exist due to multiple challenges. Solving this under-constrained optimization problem with sparse data, like a single 2D image, requires strong priors on the image formation model. 
Volumetric generative models of faces learnt from large image datasets \cite{Chan2021,gram,Karras2019stylegan2,chanmonteiro2020piGAN,buhler2023preface,tewari2022d3d} have been successful in modeling high-frequency detail including skin pores and strand-level semi-transparent hair and provide a rich latent space capable of modeling any arbitrary novel identity. They also enable semantic editing such as adding accessories (like glasses) or adding or removing wrinkles etc. However, these methods do not natively allow for relighting, which is required for accurately compositing the face into different backgrounds or environments. Some extensions \cite{deng2023lumigan,nerffacelighting,neural-3d-relightable,pan2021shadegan,pan2022gan2x} aim to disentangle the geometry and reflectance of the face from the environmental lighting by implicitly learning a subspace of intrinsic components like albedo, specularity, and normals, but, do not model the light transport accurately enough with ground truth disentangled data. In-the-wild images have a low dynamic range, non-linear photometric effects due to saturation and colored lighting, and different camera response curves. Hence, in-the-wild images are incapable of disentangling the accurate dynamic range of face reflectance, leading to dampened and inaccurate relighting results. 
However, these methods lack groundtruth supervision during training, thus, despite following physically-based rendering principles, they are physically inaccurate. This is indicated in pink in \cref{tab:related_works}.


Specialized hardware with controlled lighting, such as lightstages \cite{debevec2000acquiring}, have been used for physically accurate disentanglement of geometry and reflectance. Particularly, 2D image-based rendering (IBR) using such data has been efficiently used to achieve physically-accurate relighting of portrait images \cite{Meka:2019,yeh2022learning,Pandey21}. Several methods have attempted to learn a 3D generative model of the face geometry and reflectance using lightstage datasets. 
Therefore, the inductive bias of the trained networks of the above methods is learnt from a physically accurate dataset.
NeLF~\cite{sun2021nelf} relies on synthetically rendered ``virtual” lightstage images for training, but at inference time suffers from domain gap issues with real data, and still requires at least 5 input views captured for the same time frame, making it impractical for casual capture applications.
VoLux-GAN ~\cite{tan2022voluxgan} and VoRF~\cite{prao2022vorf, prao2023vorf} use real lightstage datasets to learn a 3D generative model of faces that can be relit under target environment maps. While VoRF does so by decomposing the input image into an ``OLAT'' reflectance basis, VoLux-GAN decomposes the image into intrinsic components and renders ``shading'' images under a target environment map that are passed through a neural rendering network to generate the relit outputs. While both methods enable photorealistic and consistent view-synthesis and physically accurate relighting, they have limited generative capacity to convert in-the-wild input images into a reflectance basis, and are weighed down at test time by the additional step of inversion and finetuning \cite{roich2021pivotal} for the given image, thus, preventing their application for interactive use cases. To circumvent the additional inversion step, several fast feedforward approaches have been proposed in the generative models literature, that train an encoder to directly predict the latent code or features given an input image \cite{richardson2021encoding,trevithick2023}, but hasn’t been extended to the relighting task. 

We propose a novel technique \OurMethod that takes an in-the-wild portrait image or video and synthesizes 3D-consistent head poses, physically plausible light editing at interactive frame rates as outlined in \cref{tab:related_works}. Our method uses a lightstage dataset to extend the generative capabilities of EG3D~\cite{Chan2021} and learns an efficient volumetric representation to implicitly disentangle face reflectance and perform relighting given a target HDRI environment map. We represent the input image as a combination of a low-dimensional latent vector and a feature image that lies in the latent manifold of EG3D by using a pre-trained geometry-aware encoder and feature alignment module by following the GAN inversion process~\cite{goae2023}. To achieve relighting, we design a simple MLP network that transforms the input latent vector to the desired illumination space by conditioning the network on a target environment map. We utilize the lighstage data~\cite{Weyrich2006Analysis} to synthesize an illumination dataset and embed it in the EG3D latent manifold to enable ground truth supervision of our pipeline. We train the relighting network with the illumination dataset to transform the inverted latent code into the desired illumination space.  Since it is challenging to encode all 3D information in the low-dimension relit latent code, similar to ~\cite{feat_inversion2022}, we refine the generator convolution layer with the combination of inverted and relit feature codes. This allows our method to learn physically accurate relighting in a 3D consistent manner, thus enabling rendering a given portrait from a novel viewpoint and performing various semantic edits made possible by the EG3D latent space. In summary, we present:

\begin{table*}[h]
\begin{center}
\caption{Our approach achieves a mix of novel capabilities for monocular in-the-wild portrait image editing such as 3D consistent pose synthesis, physically accurate relighting, semantic editing, and the efficiency of a feedforward encoder-based inference pipeline that enables interactive performance without the complexity of optimization-based fitting or finetuning. Note that while some methods like LumiGAN~\cite{deng2023lumigan} and NeRFFaceLighting~\cite{nerffacelighting} \textit{can} perform relighting, they are not physically accurate due to the low-dynamic range of in-the-wild training data.}
\begin{tabular}{|c|c|c|c|c|c|}
\hline
          & Monocular & 3D Consistency & Physical Relighting & Optimization-free & Semantic Editing \\ \hline
EG3D~\cite{Chan2021}  & \greencheck               & \greencheck           & \crossred               & \crossred    & \greencheck     \\ \hline
Live 3D Portrait~\cite{trevithick2023}  & \greencheck             & \greencheck            & \crossred               & \greencheck   & \crossred     \\ \hline
PhotoApp~\cite{mallikarjun2021photoapp}  & \greencheck            & \crossred            & \greencheck               & \greencheck   & \crossred     \\ \hline
LumiGAN~\cite{deng2023lumigan}   & \greencheck               & \greencheck           & \crossorange               & \crossred & \greencheck       \\ \hline
NeRFFaceLighting~\cite{nerffacelighting}   & \greencheck             & \greencheck           & \crossorange               & \crossred & \greencheck       \\ \hline
NeLF~\cite{sun2021nelf} & \crossred               & \greencheck           & \greencheck              & \greencheck    & \crossred   \\ \hline
VoRF~\cite{prao2022vorf, prao2023vorf}      & \greencheck              & \greencheck           & \greencheck              & \crossred     & \crossred   \\ \hline
Ours      & \greencheck            & \greencheck           & \greencheck              & \greencheck    & \greencheck   \\ \hline
\end{tabular}
\end{center}
\label{tab:related_works}
\end{table*}

\begin{itemize}
   \item A lightweight technique that enforces a strong face geometry and reflectance prior to lift 2D images to a relightable 3D space. We achieve this by leverging a pre-trained 3D generative model of faces in combination with a lightstage capture dataset to obtain a generalizable prior. 
    \item Demonstration of view synthesis and light editing of human faces from a single portrait image using the proposed prior at interactive frame rates.
    \item Extensive quantitative and qualitative evaluation of the proposed method against state-of-the-art techniques to demonstrate its enhanced efficacy.
\end{itemize}
Code and pre-trained checkpoints is available under 
\textcolor{CarnationPink}{\url{https://vcai.mpi-inf.mpg.de/projects/Lite2Relight//}}.

\section{Related Works}


We first discuss face reflectance modeling methods that are constrained by a 2D prior and/or incomplete face modeling. We then discuss 3D neural representations and facial editing methods that utilize a 3D generative model. Finally, we discuss 3D portrait lighting methods, which are the most relevant to our work. 

\paragraph{Face Appearance Modelling}
Capturing and modeling human faces to achieve highly authentic digital faces has been an active area of research \cite{debevec2000acquiring, Weyrich2006Analysis, sota2018}.
Several recent learning-based methods have exploited 2D generative image models for facial relighting \cite{tewari2020pie,stylerig,mallikarjun2021photoapp,abdal2020styleflow,richardson2021encoding,kwak2022injecting}.
However, they cannot consistently disentangle the underlying identity-specific geometry from the view-dependent appearance, leading to inconsistent view synthesis. 
Parametric face models \cite{blanz1999morphable,flame} have traditionally provided 3D priors for such tasks, but suffer from their low-dimensional representations, which limit their capacity to model high-frequency details such as wrinkles, and completely fail for unstructured regions like hair.  
Although there exist methods that accurately capture and model face reflectance fields, rendering such digital avatars~\cite{emily2010, seymour2017} requires significant manual effort. Several traditional method use hand-crafted models and target specific parts of the face, such as facial hair~\cite{hair}, skin wrinkles~\cite{gotardo2018}, eyes~\cite{eyenerf} teeth~\cite{wu2016teeth} and lips \cite{garrido2016lips} using computationally expensive optimization routines, and often require a dense and invasive data capture mechanisms. 
Using a parametric face model, multiple methods \cite{mbr_frf,shugo2018} enable face reflectance editing in the face interior region for monocular inputs. 
Several image-based relighting methods \cite{Meka19, nestmeyer2020faceRelighting, Zhou_2019_ICCV, Pandey21, sipr_sun, sipr_ex} relight entire human heads for a fixed viewpoint or identity-specific settings \cite{Bi21}. Due to the lack of an underlying 3D representation such methods are limited to only relighting as they cannot modify the camera viewpoints.

\paragraph{3D Neural Representations and GANs}
Neural Radiance Fields (NeRF) \cite{mildenhall2020nerf}, model a 3D scene as a 5D continuous radiance field function using a multi-layer perceptron (MLP) network and differential volume rendering from multiple viewpoints. This innovative approach enables precise 3D representations without the need for explicit geometric modeling.
In the realm of 3D neural rendering, NeRF-based methods \cite{Tewari2022NeuRendSTAR} have successfully achieved realistic rendering of human avatars \cite{Gafni_2021_CVPR, teotia2023hq3davatar} in a consistent 3D manner. Several innovative approaches \cite{iphoneavatar, headnerf, ramon2021h3d, Buhler_2023_ICCV, rome2023} have expanded these techniques into multi-identity models, learning a facial prior and demonstrating personalization even with sparse input data.

There has been a significant effort \cite{Chan2021, orel2022styleSDF, gu2021stylenerf, gram, Chan2022} to blend Generative Adversarial Networks (GANs) \cite{goodfellow2014generative} with NeRF \cite{mildenhall2020nerf}, facilitating the learning of a latent facial manifold. Notably, EG3D \cite{Chan2022} incorporates StyleGAN \cite{Karras2019stylegan2} into a 3D framework to generate a comprehensive generative 3D prior of faces.
These rich generative priors have opened avenues for portrait editing, primarily utilizing GAN inversion techniques. Methods such as \citet{richardson2021encoding, feat_inversion2022, roich2021pivotal} have demonstrated embedding portrait images into StyleGAN's latent space. In 3D GANs, especially EG3D, optimization-based inversion methods \cite{ko20233d, xie2022high} have been used to update inverted latent codes to minimize reconstruction loss. However, such methods can be slow and often yield subpar editing quality \cite{feat_inversion2022}, as the optimized latent code could diverge from the original sampling space. Encoder-based inversion methods \cite{live3dinversion, goae2023}, in contrast, offer faster performance with better regularization due to the lack of an optimization loop. Live 3D Portrait \cite{live3dinversion} uses a ViT-based architecture \cite{dosovitskiy2020vit} to learn a new triplane representation \cite{Chan2022} using synthetic data from EG3D, efficiently converting 2D portraits to 3D while foregoing the rich latent manifold for semantic editing.

Drawing inspiration from E4E \cite{richardson2021encoding}, GOAE \cite{goae2023} trains an encoder to embed subjects within EG3D's $W^+$ space, also incorporating an attention-based module to recover identity-specific features. This approach allows inverted portraits to be elevated to 3D while retaining the capability for semantic editing. We adapt such a 3D-aware encoder to invert desired portraits into the latent space of EG3D and perform relighting at interactive rates.

\paragraph{3D Portrait Relighting}
%
Volumetric rendering approaches have enabled simultaneous editing of viewpoints and illuminations for both general scenes and human avatars without requiring explicit 3D geometry. The work of \citet{rudnev2022nerfosr, boss2021nerd, zhang2021nerfactor} perform intrinsic decomposition of general scenes and relight under novel illumination. Using a lightstage setup, \citet{yang2023light, sarkar2023litnerf, tr_avatar} demonstrate person-specific relighting. 
Using a synthetic OLAT dataset~\citet{sun2021nelf} adapt PixelNeRF~\cite{yu2020pixelnerf} to learn a generalizable 3D portrait relighting method. Similarly, MEGANE~\cite{megane} trains an MVP~\cite{Lombardi21} representation that can generalize to unseen subjects. Both methods need at least three multi-view inputs, which limits their application in many real-world scenarios. Instead, our method takes a single monocular image as input.

Using a similar lightstage dataset, VoRF~\cite{prao2022vorf, prao2023vorf} trains a NeRF-based autodecoder network that generalizes to unseen identities under monocular settings. However, these models, trained on data that was captured in controlled setups with limited numbers of subjects often lead to a less diverse face prior, resulting in limited generalization towards in-the-wild samples. The method we present here not only avoids these issues using the EG3D prior but also computes results at interactive rates because it does not rely on a rather costly implicit representation.
Generative models like EG3D offer a rich face prior and can synthesize arbitrary numbers of faces. On this basis, recent methods \cite{tan2022voluxgan, neural-3d-relightable, deng2023lumigan} relight synthetic identities sampled from a latent space. Here, \citet{deng2023lumigan} combine precomputed radiance transfer \cite{Sloan2002PrecomputedRT} with adversarial learning to relight portraits, but due to the self-supervised learning paradigm this method struggles to learn physically accurate lighting. Additionally, the above methods focus on generating synthetic samples and offer very limited capability for editing a given input portrait. Our method on the other hand can be controlled very accurately by an explicit environment map as input.

\begin{figure*}
\centering
\newcommand{\mywidth}{1.0}
\includegraphics[]{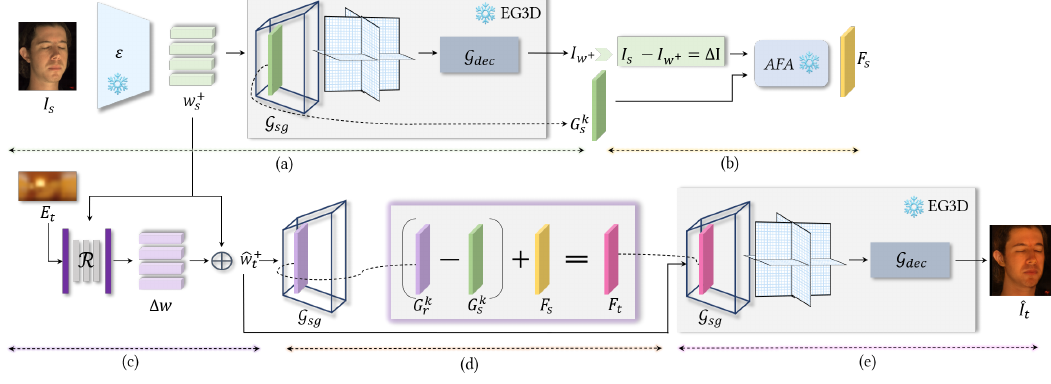}
\caption{\textbf{Method Overview.} (a) Given an input image $\srcImg$, we use a pretrained encoder $\encoder$ to invert $\srcImg$ and obtain the latent vector $\srclatw$. We pass $\srclatw$ through a pretrained EG3D network to render the inverted image $\invImg$ and extract convolutional features $\srcgenFeat$ from $\gen$. (b) Next, we use image residual $\diffImg$ and $\srcgenFeat$ as inputs to the AFA module, to obtain $\srcrecFeat$. (c) Given a target environment map $\tgtemap$, our relighting network $\latentNet$ generates $\deltalatw$, which is combined with $\srclatw$ to produce the relit latent code $\relitlatw$. (d) Subsequently, we obtain $\relitrecFeat$ by following \cref{eq:feat_manipulate}. (e) Finally, we replace the $k$-th convolutional feature of $\gen$ by $\relitrecFeat$ and perform a full forward pass through the EG3D network with the latent code  $\relitlatw$ to generate $\relitrecImg$, which is relit by $\tgtemap$. Note: $\decoder$ takes camera pose $\pose$ as input.}
\label{fig:overview}
\end{figure*}
\section{Method}

The primary aim of our method is to relight a portrait of a human from a single input image under any desired novel viewpoint and illumination. This is achieved without the need for time-intensive optimization processes. This task is inherently underconstrained due to depth ambiguity and the complex interplay between facial features and varying illumination. Directly modeling light transport is computationally expensive and approximations often lead to non-photorealistic outcomes. Our approach circumvents these challenges by implicitly managing light transport through neural networks.

As illustrated in \cref{fig:overview}, we leverage EG3D \cite{Chan2022}, a 3D-aware generative model known for its high-quality, generalizable representations of human faces. To adapt the relighting task to the feature space of EG3D, we first embed real images into the EG3D space \cref{sec:prior_model}. Following this, a mapping network, trained on a lightstage dataset (\cref{sec:data}), is employed to transition the original feature vector into a target feature vector. This enables us to render the face under novel lighting conditions and viewpoints. Details of the relighting module are further elaborated in \cref{sec:method_relight}. Finally, the various loss functions utilized in our method are outlined in \cref{sec:obj_function}.

\subsection{Dataset}\label{sec:data}
We use the multi-view lightstage dataset captured by \cite{Weyrich2006Analysis} that has 353 subjects illuminated under $N=150$ point light sources where each subject is captured with 16 cameras. Hence, the dataset contains a set of one-light-at-a-time (OLAT) images $O = \{\imageolat_1, ... \imageolat_N | \imageolat_i \in \setR^{512 \times 512 \times 3}\}$ for every subject under 16 viewpoints. Normally, ``in the wild'' images are not captured under such OLAT conditions. Therefore, directly embedding OLAT images onto the EG3D space will result in unfaithful reconstructions. Thus, to alleviate this and obtain realistic scene illumination conditions, we linearly combine OLAT images with environment maps by following image-based relighting ~\cite{debevec2000acquiring} as follows 
%
%
%
\begin{equation}\label{eq:relit}
      \natImg  = \sum_{\incid = 0}^{N} \factor(\incid) \cdot \imageolat_{\incid}
\end{equation}
where $\natImg \in \setR^{512 \times 512 \times 3}$  is the relit image and $\factor : \setN_{< N} \rightarrow \setR^3$ is the downsampled version of the input environment map.

\paragraph{Training Data} We relight all the subjects under 50 natural illumination conditions randomly sampled from the Laval Indoor and Outdoor datasets \cite{holdgeoffroy-cvpr-19,gardner2017learning} by using \cref{eq:relit}. To obtain paired data for supervision, for each training step we randomly sample two naturally relit images of the \textit{same subject}: The input i.e. the source image is referred to as $\srcImg$ and is relit with a source environment map $\srcemap$. The target $\tgtImg$ is relit with $\tgtemap$. 


\subsection{3D GAN Inversion}
\label{sec:prior_model}
 We adopt EG3D~\cite{Chan2022}, a 3D-aware generative model that has demonstrated remarkable generalization results, as our backbone. The generator $\gen$, based on StyleGAN \cite{Karras2019stylegan2}, maps a latent vector $\srclatw$ and camera pose $\pose$ to triplane features that are further decoded to render a low-resolution image. This low-resolution image is  upsampled to $512 \times 512$ using a super-resolution module.  We formally denote the decoder, volume rendering, and upsampling as $\decoder$. The generated image is obtained as:
\begin{equation}\label{eq:decoder_eg3d}
        \invImg = \decoder(\gen(\srclatw), \pose)
\end{equation}

To obtain robust features in the EG3D space that are representative of the training images of our dataset, we use an encoder $\encoder$, adopting \cite{goae2023}.  $\encoder$ maps a given source image $\srcImg$ to a latent code $\srclatw \in \setR^{14 \times 512}$: 
\begin{equation}\label{eq:inversion}
        \srclatw = \encoder(\srcImg) 
\end{equation}
However, $\srclatw$ is a low-dimensional latent code that is insufficient to learn a rich representation of the portrait. Thus, we follow \cite{goae2023} and further adapt their ``Adaptive Feature Alignment'' module which has attention-based and convolutional layers to learn an additional feature code. We denote the module as $\afa$ and it takes the difference between the source image and the predicted image as input, i.e. $\diffImg = \srcImg - \invImg$, along with features of $\gen$:
\begin{equation}\label{eq:inversion-afa}
        \srcrecFeat = \afa(\diffImg, \srcgenFeat) \in \setR^{32 \times 32 \times 512}
\end{equation}
where $\srcgenFeat \in \setR^{32 \times 32 \times 512}$ is the $\rk$-th convolutional feature of $\gen$, for some fixed $\rk$. 
Finally, we use the remaining layers of $\gen$, starting from $\rk+1$, to obtain the triplane features that are used to reconstruct the input image. Thus, $\srclatw$ and $\srcrecFeat$ together serve as a robust representation of $\srcImg$ in a high-dimensional latent space.

\subsection{Relighting}\label{sec:method_relight}
In this section, we explain the steps involved in relighting $\srcImg$ given a target environment map $\tgtemap$. The latent space of EG3D contains a rich representation of faces and we leverage this advantageous latent space to perform relighting. Our aim is to embed the target images $\tgtImg$ to this latent space and obtain $\relitlatw \in \setR^{14 \times 512}$. However, as described in \cref{sec:data}, a pair of $\srcImg$ and $\tgtImg$ have the same subject relit under different environment maps. Thus, we have an MLP $\latentNet$, that maps the source latent code $\srclatw$ and the target environment map $\tgtemap$ to a latent offset $\deltalatw$:
\begin{equation}\label{eq:relight_mlp}
    \deltalatw = \latentNet(\srclatw, \tgtemap)  \in \setR^{14 \times 512}
\end{equation}
To obtain $\relitlatw$, we add this offset to $\srclatw$:
\begin{equation}\label{eq:offsets_adding}
    \relitlatw = \srclatw + \deltalatw 
\end{equation}
A relit image can be directly obtained by giving $\relitlatw$ as the input to $\decoder$. However, we found the obtained image to not preserve subject-specific details well, altering the perceived identity (see ablation in \cref{abln_feature_code}). This is because $\relitlatw$ is a low-dimensional representation that is insufficient to capture the subject details. But, the $\afa$ model can not be used as it requires the relit image which we don't have access to. Hence, we aim to transfer fine-scale details present in $\srcrecFeat$ to the target illumination space: We perform a forward pass through $\gen$ with $\relitlatw$ as input and extract the $\rk$-th convolutional feature $\relitgenFeat$. Finally, we combine the convolutional features of the source and target image by following \cite{feat_inversion2022}: 
\begin{equation}~\label{eq:feat_manipulate}
    \relitrecFeat := \srcrecFeat + \relitgenFeat - \srcgenFeat \in \setR^{32 \times 32 \times 512}
\end{equation}
 Thus, $\relitlatw$ and $\relitrecFeat$ serve as a robust representation of the predicted relit image in a high-dimensional latent space. Finally, we replace the $\rk$-th convolutional feature of $\gen$ with  $\relitrecFeat$ and perform a full forward pass as follows:
\begin{equation}
        \relitrecImg = \decoder(\gen(\relitlatw, \relitrecFeat), \pose) \\
\end{equation}

\subsection{Loss Functions}~\label{sec:obj_function}
We train $\latentNet$, while the networks $\gen$, $\decoder$ and $\afa$ remain frozen.
As described in \cref{sec:data}, we first obtain a set of paired source and target images $\srcImg$, $\tgtImg$.
Given $\srcImg$ as input and target illumination condition $\tgtemap$, the goal is to predict a $\relitrecImg$ that is as close to $\tgtImg$ as possible.
We obtain target latent code $\tgtlatw$ as described in Eq~\ref{eq:inversion} using $\tgtImg$.
As a training objective, we minimize a combination of reconstruction and latent loss:

\paragraph{Reconstruction Loss}:
%
We penalize deviations of $\relitrecImg$ from $\tgtImg$ by $L1$ distance:
\begin{equation}
    \lossc = \Norms{\relitrecImg - \tgtImg}
\end{equation}

\paragraph{Perceptual Loss}:
Supervision in the image space alone resulted in poor reconstruction of certain illumination conditions in the target space. Therefore, we employ a feature-based loss $\lossp$~\cite{justin_percetptual_loss} between $\relitrecImg$ and $\tgtImg$. 
\begin{equation}
    \lossp = \Norm{\Phi_{vgg}(\relitrecImg) - \Phi_{vgg}(\tgtImg)}
\end{equation}
Where $\Phi_{vgg}$ is the extracted features from the pre-trained VGG~\cite{simonyan2015deep} network.  We conduct an ablative study (\cref{abln_subsec_loss}), to demonstrate the effectiveness of $\lossp$ for relighting.

\paragraph{Latent Loss}: 
To ensure that the $\relitlatw$ predicted by $\latentNet$ is in the same part of the EG3D latent space as $\srclatw$ (and not in a region that behaves differently under $\gen$),  we penalize the $L2$ distance between $\relitlatw$ and $\tgtlatw$ :
\begin{equation}
    \lossl = \Norm{\relitlatw - \tgtlatw}
\end{equation}


The total loss is given as:
\begin{equation}
    \lossall = \wtl\lossl + \wtc\lossc + \wtp\lossp 
\end{equation}

\section{Evaluation}
We describe the datasets used for evaluation in \cref{subsec:eval_data}. In \cref{subsec:results_relighting} we discuss qualitative evaluation of simultaneous view synthesis and relighting for in-the-wild portraits. We report quantitative analysis on the lightstage dataset in \cref{subsec:results_baseline_compare} and discuss ablative experiments in \cref{subsec:results_ablation}.

\subsection{Dataset}\label{subsec:eval_data}
We create an evaluation dataset based on a lightstage dataset \cite{Weyrich2006Analysis} with 10 unseen subjects, illuminated under 10 novel illumination conditions under 12 novel camera viewpoints. 
Linearly combining OLATs with downsampled HDRI environment maps (like in \cref{sec:data}), gives us relit ground truth images.
Additionally, we qualitatively evaluate \OurMethod using diverse subjects captured in the wild~\cite{ravdess, Shih2014, ramon2021h3d, caselles2023implicit}.


\subsection{Relighting in-the-wild Portraits} \label{subsec:results_relighting}
\input{tex_fig/results/in_the_wild_main}
In  \cref{fig:ours_in_the_wild} and \cref{fig:ours_in_the_wild_additional}, we show \OurMethod{}'s capability to modify viewpoint and illumination of diverse in-the-wild portraits. Our method retains 3D consistency for head poses as well as relighting. This can be specifically observed in the last two columns of both figures, where the subject is relit under the same environment map, and shows consistent relighting under two different camera views. This is because the 3D representation, coupled with accurate relighting in the latent manifold, faithfully preserves the identity and expression of subjects.
From (\cref{fig:ours_in_the_wild} and \cref{fig:ours_in_the_wild_additional}) we observe strong identity preservation. Moreover, intricate details like expressions (see rows 1 of \cref{fig:ours_in_the_wild} and \cref{fig:ours_in_the_wild_additional}) and accessories such as spectacles under varying illumination scenarios (see row 4 of \cref{fig:ours_in_the_wild_additional}) are preserved as well.
Owing to the robust generative prior, our method generalizes to a wide variety of subjects, while preserving/synthesizing complex reflectance phenomena, such as subsurface scattering and specular highlights, particularly noticeable on the nose, cheeks, and forehead (see rows 2 and 3 of \cref{fig:ours_in_the_wild}), as well as soft-shadows (see row 1 of \cref{fig:ours_in_the_wild}, rows 2 and 3 of \cref{fig:ours_in_the_wild_additional}).
\subsection{Comparisons to Previous Works} \label{subsec:results_baseline_compare}
We compare our method to state-of-the-art methods for simultaneous viewpoint and illumination editing: 
(1) VoRF  \cite{prao2022vorf, prao2023vorf} employs a NeRF-based auto-decoder architecture learning a volumetric reflectance field from an OLAT lightstage dataset. 
(2) PhotoApp \cite{mallikarjun2021photoapp} leverages a 2D StyleGAN prior for faces, focusing on learning latent transformations to manipulate viewpoint and illumination. 
(3) NeRFFaceLighting \cite{nerffacelighting} leverages EG3D design principles to learn a separate appearance and lighting triplane. 
(4) NeLF \cite{sun2021nelf} trains pixelNeRF\cite{yu2020pixelnerf} inspired 3D representation using a synthetic lightstage dataset to derive 3D geometry and facial reflectance properties. 
For all the methods, we use the original implementation provided by the authors and train it on the lightstage dataset. For a fair qualitative assessment, VoRF results are upsampled to $512 \times 512$ pixels. \cref{fig:baseline-vorf-pa,tab:baseline-single_view} indicate that \OurMethod not only surpasses the baseline approaches in rendering high-quality relit images under various scene illuminations but also excels in maintaining 3D consistency and identity fidelity.

\begin{table}[ht]
\centering
\caption{\textbf{Quantitative Results: Comparisons to Previous Works}. We report SSIM, landmarks distance (LD), and PSNR computed on the the test data (\cref{subsec:eval_data}), for \OurMethod and previous works, relighting subjects under novel views.}
\begin{tabular}{lcccc}
\hline
 & SSIM $\uparrow$ & LD $\downarrow$ & PSNR $\uparrow$ \\ \hline
NeLF (3-views) & 0.75 & NA & 19.72 \\
PhotoApp & 0.72 & 34.08 & \textbf{29.13} \\
VoRF  & 0.69 & 16.90 & 20.21 \\
NeRFFaceLighting  & 0.79 & 28.31 & 13.41 \\\hline
\OurMethod &  \textbf{0.83} & \textbf{9.76} & 28.3 \\ \hline
\end{tabular}
\label{tab:baseline-single_view}
\end{table}


\paragraph{Metrics} Apart from Peak Signal-Noise Ratio (PSNR) and Structural Similarity Index Measure (SSIM), we also use a landmark distance metric (LD) to evaluate the 3D consistency of facial geometry. It is calculated as the average deviation of 68 facial key points  \cite{bulat2017far} over the evaluation dataset.

\paragraph{VoRF} \cref{fig:ours_vs_vorf} shows that VoRF is unable to generalize to subjects that are out of the training distribution (lightstage data): It struggles with identity preservation (row 1), as well as expression preservation and capturing eye details (rows 2 and 3). This is due to VoRF's dependence on a face prior learned from the lightstage data. For lightstage subjects  \cref{fig:baseline-vorf-pa} VoRF does show reasonable results.  In contrast, however, \OurMethod can preserve facial details, expressions, and even eye details for out-of-distribution data (columns 4 and 5 in  \cref{fig:ours_vs_vorf}), thanks to our design where we leverage information from large-scale in-the-wild data (FFHQ). Furthermore, due to VoRF's NeRF-based framework, which uses large amounts of memory, its outputs are confined to resolution $128 \times 128$ (see blurriness in results in  \cref{fig:baseline-vorf-pa}). \OurMethod, on the other hand, based on the triplane framework,  can preserve fine details at resolution  $512 \times 512$. Finally, VoRF requires a two-step optimization process that takes $\sim$ 10mins to relight a single image, making it unsuitable for interactive applications.
While \OurMethod, with its encoder-based architecture and an efficient triplane-based volume rendering technique, achieves interactive performance (7-31 fps. See Sec. 1 of SupMat).

\paragraph{PhotoApp} PhotoApp achieves high image quality (see PSNR in \cref{tab:baseline-single_view}), thanks to the latent face prior from StyleGAN \cite{Karras2019stylegan2}. However, PhotoApp's inherent lack of a native 3D representation leads to noticeable inconsistencies in subject identity under novel views. This results in much lower SSIM and LD scores compared to \OurMethod{}, which uses an efficient NeRF-based representation, and excels in maintaining 3D consistency. \cref{fig:baseline-vorf-pa} visualises this advantage: PhotoApp deviates from the original identity, while \OurMethod maintains a close resemblance to the input subject.

Furthermore, Photoapp manipulates both viewpoint and illumination in the latent space which is challenging to control (row 2 in  \cref{fig:baseline-vorf-pa}). Lite2Relights's latent transformation is confined to illumination, allowing us to control the
viewpoint and illumination in a disentangled way. Further results shown in \cref{fig:ours_vs_photoapp} demonstrate that PhotoApp struggles to preserve the identity (rows 1 and 3) and even shows inconsistent illumination across different views. \OurMethod does preserve the identity of the subject in a 3D-consistent manner and maintains the desired lighting across all views.

\paragraph{NeRFFaceLighting} In our analysis of NeRFFaceLighting (NFL), as illustrated in \cref{fig:baseline-vorf-pa}, NFL clearly shows significant lighting artifacts, poor identity retention (can be noticed in row 1), and overly saturated effects (observed in row 2). These shortcomings are quantitatively evidenced in \cref{tab:baseline-single_view}, where NFL consistently underperforms in comparison to \OurMethod.
We hypothesize that these issues stem from the inherent complexities in jointly optimizing illumination and identity during the inversion process. NFL employs an explicit decomposition of the input into albedo and shading components, which makes the inversion process complicated and fails to recover the true identity accurately. 
As a result, the input light gets baked into the albedo during relighting, as shown in row 2, where the prediction of the subject has a red-colored skin taken from the input light source. 
In contrast, our method mitigates the albedo-lighting ambiguity by utilizing a supervised training approach with a lightstage dataset, where each subject is relit under various lighting conditions. Moreover, our strategy includes an encoder-based inversion process that helps preserve identity integrity and achieve more accurate relighting outcomes.

\paragraph{NeLF} NeLF requires at least three input views to obtain reasonable results. This makes NeLF unusable for almost all in-the-wild portraits. Moreover, NeLF struggles to reconstruct a reasonable facial structure even with multiview inputs as it fails to represent the underlying geometry leading to low scores in \cref{tab:baseline-single_view}.
We can observe the results in \cref{fig:ours_vs_nelf}, where NeLF's results were obtained from 3 input views producing distorted facial reconstructions. In contrast, our approach demonstrates superior generalization to novel subjects from a single image. Furthermore, our method exhibits the ability to accurately relight these subjects, maintaining both the integrity of facial features and the overall photorealism.
\subsection{Ablation Study} \label{subsec:results_ablation}
\paragraph{Significance of Feature Code Manipulation} \label{abln_feature_code}
To evaluate the importance of \cref{eq:feat_manipulate}, particularly focusing on the comparative roles of $\relitrecFeat$ and $\relitgenFeat$, we compare our original method to a variant that does not replace $\relitgenFeat$ by $\relitrecFeat$.  The results, as depicted in \cref{fig:abln_feat_code}, reveal that the incorporation of $\relitrecFeat$ significantly enhances the recovery of fine-scale, identity-specific details. This improvement is especially noticeable in the eye region of row 2, as well as in the jawlines and overall facial contours of all the rows. The efficacy of $\relitrecFeat$ in enhancing the fidelity of these features is further corroborated by its superior performance in quantitative evaluations, see \cref{tab:abln_lpips_feat}.
\begin{table}[ht]
\centering
\caption{\textbf{Quantitative Results: Ablation Study}. We report SSIM, landmarks distance (LD), and PSNR on the test data of lightstage, where  subjects are relit under novel viewpoints.} 
\begin{tabular}{lcccc}
\hline
           & SSIM $\uparrow$ & LD $\downarrow$ & PSNR $\uparrow$ \\ \hline
w/o F-space  & 0.831 & 10.44  & 28.29 \\
w/o $\lossp$ & 0.830 & 10.12 & 28.32 \\ \hline
Full Model &  \textbf{0.83} & \textbf{9.76} &\textbf{28.33} \\ \hline
\end{tabular}
\label{tab:abln_lpips_feat}
\end{table}
\paragraph{Importance of Perceptual Loss}\label{abln_subsec_loss}
It is feasible to supervise $\latentNet$ solely using $\lossl$ and $\lossc$, omitting $\lossp$, thereby effectively only supervising the latent vectors to preserve identity. However, as our findings in \cref{fig:abln_loss_lpips} suggest, relying solely on reconstruction losses may not guarantee accurate relighting: Rows 1 and 2 show inaccuracies in the overall color of the images, while shadows and highlights are noticeably absent in row 3. We hypothesize that these discrepancies arise due to the inherent limitations of reconstruction losses, which do not necessarily account for perceptual plausibility. Adding $\lossp$, as suggested before \cite{justin_percetptual_loss}, ensures a more accurate relighting, as demonstrated in the column labeled ``{w/ $\lossp$}''. Here, the relit images not only more closely resemble the reference illumination but also achieve higher scores across quantitative metrics.
Additional ablations are provided in the SupMat. 
\section{Conclusion}
In conclusion, our work, \OurMethod, represents a significant advancement in the field of 3D portrait editing and relighting, effectively addressing the complex challenges associated with interactive, photorealistic image processing for AR/VR applications. By innovatively leveraging a pre-trained 3D generative model in conjunction with a lightstage dataset and a 3D-aware encoder, we have successfully developed a method that lifts 2D images into a relightable 3D space with strong geometric and reflectance accuracy. This technique not only simplifies the process of achieving photorealism in portrait editing but also ensures practical applicability by minimizing the need for extensive computational resources and complex data inputs.
Our work with \OurMethod showcases the capability of performing view synthesis, light editing, and semantic modifications at interactive rates from single, in-the-wild portrait images, a notable development in the field. The efficiency and fidelity with which \OurMethod accomplishes these tasks have meaningful implications for the evolution of personalized digital content creation, particularly within the context of AR/VR applications.
The extensive evaluations of our method against current state-of-the-art techniques highlight its superior performance and practicality, reinforcing our contribution to the domain of computational photography and graphics. We believe that the release of our code and pretrained models will foster further research and development, potentially leading to widespread adoption and continuous improvement in this field.
\begin{acks}
This work was supported by the ERC Consolidator Grant 4DReply (770784). We extend our gratitude to Shrisha Bharadwaj for providing feedback and constant support.
\end{acks}

\bibliographystyle{ACM-Reference-Format}
\bibliography{main.bib}

\pagebreak
\input{tex_fig/results/ls_compare}
\input{tex_fig/results/in_the_wild_additional}
\clearpage
\phantom{This is a fix for image rendering on the last page.}
\input{tex_fig/results/compare_photoapp}
\input{tex_fig/results/compare_nelf}
\input{tex_fig/results/compare_vorf}
\input{tex_fig/results/abln_feat_code}
\input{tex_fig/results/abln_loss_lpips}

\end{document}